%% file: main.tex
\documentclass[cameraready]{Interspeech}

\title{Adapting Text LLMs to Speech via Multimodal Depth Up-Scaling}

\author[affiliation={1}]{Kazuki}{Yano}
\author[affiliation={1}]{Jun}{Suzuki}
\author[affiliation={1,2}]{Shinji}{Watanabe}

\address{
    $^1$ Tohoku University, Japan \\
    $^2$ Carnegie Mellon University, USA
}

\email{yano.kazuki@dc.tohoku.ac.jp}

\keywords{Speech Language Model, Speech Recognition}

\usepackage{comment}
\usepackage{multirow}
\usepackage{cite}
\usepackage{silence}
\begin{document}

\maketitle

\begin{abstract}
Adapting pre-trained text Large Language Models (LLMs) into Speech Language Models (Speech LMs) via continual pre-training on speech data is promising, but often degrades the original text capabilities.
We propose \textit{Multimodal Depth Up-scaling}, an extension of an emerging strategy in continual LLM pre-training, where new transformer layers are inserted into a frozen text LLM and only the added layers are trained on speech data.
Experiments with SmolLM2-360M and SmolLM2-1.7B on 48k hours of English Automatic Speech Recognition (ASR) data show that depth up-scaling achieves ASR comparable to full fine-tuning while causing far less text degradation than both full fine-tuning and Low-Rank Adaptation (LoRA).
We further show that incorporating E-Branchformer, an architecture designed for speech recognition, as the inserted layers achieves ASR that matches or surpasses full fine-tuning on the larger model while reducing text degradation by over 75\% with 60\% fewer trainable parameters.
\end{abstract}

\section{Introduction}\label{sec:intro}

Pre-trained large language models (LLMs) possess strong language understanding and generation capabilities acquired from large-scale text corpora~\cite{brown2020language, grattafiori2024llama}.
This progress has extended to the speech domain, giving rise to Speech Language Models (Speech LMs)~\cite{tang2024salmonn,chu2024qwen2,tian25b_interspeech, arora2025on} that convert speech-related tasks into sequential modeling built on top of pre-trained text LLMs.
A common approach is to adapt a text LLM into a Speech LM through continual pre-training (CPT) on speech data, allowing the model to inherit the linguistic knowledge of the base LLM for tasks such as automatic speech recognition (ASR).

However, CPT on speech data often degrades the text capabilities acquired during LLM pre-training~\cite{tian25b_interspeech,hsiao2025analyzing}, undermining the advantage of building on a pre-trained LLM.
Existing approaches to reduce this degradation include text data replay during speech training~\cite{tian25b_interspeech, peng-etal-2025-voicetextblender, hsiao2025analyzing} and parameter-efficient fine-tuning methods such as Low-Rank Adaptation (LoRA)~\cite{hu2022lora}, which is widely adopted in Speech LMs~\cite{tang2024salmonn, hu-etal-2024-wavllm} and argued to better preserve text performance~\cite{biderman2024lora}.
However, replay requires additional text data and increases training cost proportionally. 
Moreover, most open-weight LLMs do not release their pre-training data~\cite{grattafiori2024llama,chu2024qwen2, yang2025qwen3}, making faithful replay infeasible in practice. 
LoRA, on the other hand, may not provide sufficient capacity for learning a new modality, as we show in our experiments.

In this work, we propose multimodal depth up-scaling, an extension of depth up-scaling~\cite{wu-etal-2024-llama, yano-etal-2025-step,cao2025progressive} to cross-modal adaptation, where new transformer layers are inserted into a frozen text LLM and only the added layers are trained on speech data.
Since the original weights are never modified, the added layers can be removed at inference to recover the pre-trained model exactly.

We conduct experiments using SmolLM2-360M and SmolLM2-1.7B~\cite{allal2025smollm} as base models, training on 48k hours of English ASR data from the OWSM v3.2 suite~\cite{tian2024effects}.
We show that multimodal depth up-scaling achieves ASR performance comparable to full fine-tuning while causing far less text task degradation than both full fine-tuning and LoRA.
We also investigate layer placement strategies and show that incorporating E-Branchformer~\cite{kim2023branchformer}, an architecture designed for speech recognition, as the inserted layers matches or surpasses full fine-tuning on the larger model while reducing text task degradation by over 75\% with 60\% fewer trainable parameters.
We further demonstrate that multimodal depth up-scaling retains the base LLM's ability to follow text instructions such as translation and summarization, while full fine-tuning completely loses this ability.
Our findings provide useful guidance for adapting text LLMs to the speech modality without sacrificing their original capabilities.

\begin{figure}[t]
    \centering
    \includegraphics[width=0.45\textwidth]{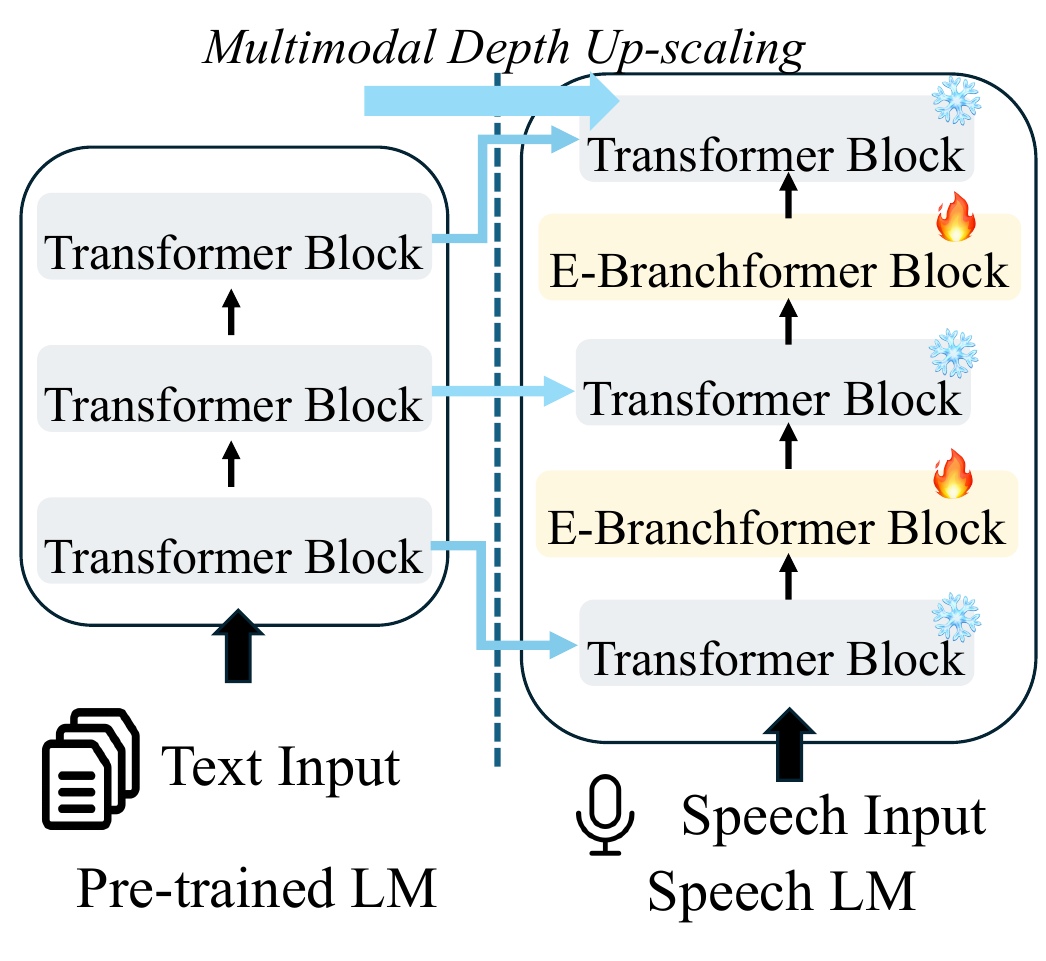}
    \caption{Proposed multimodal depth up-scaling strategy for Speech LM adaptation: trainable E-Branchformer blocks are interleaved with frozen transformer blocks to acquire speech capabilities while preserving pre-trained text representations.}
    \label{fig:method}
    \vspace*{-5mm}
\end{figure}

\section{Related Work}\label{sec:related}

\noindent\textbf{Speech LMs and Forgetting.}
Recent work has adapted pre-trained text LLMs to process speech, either by connecting a speech encoder to the LLM~\cite{tang2024salmonn,chu2024qwen2} or by using discrete speech tokens that can be directly modeled by the LLM~\cite{tian25b_interspeech}.
While these models can inherit the language ability of the base LLM, CPT on speech data often degrades the original text capabilities.
To address this, several studies replay text data during speech training to preserve text capabilities~\cite{tian25b_interspeech, peng-etal-2025-voicetextblender}, and experience replay has been explored in broader continual speech learning settings~\cite{hsiao2025analyzing, wang2025continual,javed2025nirantar}.
Parameter-efficient methods such as LoRA~\cite{hu2022lora} constrain weight updates to low-rank subspaces to reduce forgetting~\cite{biderman2024lora,hsiao2025analyzing}, and have been widely adopted in Speech LMs for cross-modal adaptation~\cite{tang2024salmonn, hu-etal-2024-wavllm}.
Depth up-scaling offers a different approach by leaving all original parameters untouched and adding new capacity through additional layers, without requiring text data during speech training.

\noindent\textbf{Depth Up-scaling.}
Depth up-scaling increases a model's capacity by adding new transformer layers.
Several studies have applied this idea to LLMs by duplicating and inserting layers for continued pre-training~\cite{kim-etal-2024-solar, wu-etal-2024-llama}, and recent work has further improved layer initialization through learnable prediction~\cite{yang-etal-2025-lesa} and optimal transport~\cite{cao2025progressive}.
These methods have been applied to text-domain expansion, but their application to multimodal speech adaptation remains unexplored.
We apply depth up-scaling to adapt a text LLM for speech with all original layers frozen, and systematically investigate how the placement and architecture of added layers affect both ASR and text task performance.

\section{Method}\label{sec:method}

Our goal is to adapt a pre-trained text LLM into a Speech LM that can perform ASR while preserving the original text capabilities, including the ability to follow text instructions at inference.
We first expand the vocabulary to handle speech tokens (\S\ref{sec:vocab}), then add new transformer layers via depth up-scaling (\S\ref{sec:depth}), as illustrated in Figure~\ref{fig:method}.
We also consider E-Branchformer as an alternative architecture for the added layers (\S\ref{sec:arch}), and describe how the added layers can be dropped at inference to fully recover the pre-trained model (\S\ref{sec:drop}).

\subsection{Vocabulary Expansion}\label{sec:vocab}

To process speech input, we expand the tokenizer vocabulary of the pre-trained LLM with speech tokens.
Following OpusLM~\cite{tian25b_interspeech}, we convert speech into discrete tokens using a combination of semantic and acoustic tokens, as described in~\cite{tian-etal-2025-espnet}.
These speech tokens are appended to the original text vocabulary.
The embedding layer is resized accordingly and trained together with the added layers, while the original embedding weights are kept frozen.

\subsection{Depth Up-scaling}\label{sec:depth}

Rather than fine-tuning the existing parameters, we insert $m$ new transformer layers into the pre-trained model that originally consists of $n$ layers $\{f_i\}_{i=1}^{n}$, as shown in Figure~\ref{fig:method}.
All original layers $\{f_i\}_{i=1}^{n}$ are frozen during training, and only the added $m$ layers $\{g_j\}_{j=1}^{m}$ are trainable.
Since the original weights are never modified, the added layers can be removed at inference to recover the pre-trained model exactly.

Prior work on model expansion has shown that the expanded model should initially behave identically to the original, a property known as function preserving initialization~\cite{wu-etal-2024-llama,chen-etal-2022-bert2bert}.
Without this property, the added layers would immediately perturb the pre-trained representations, potentially destabilizing training.
Each added layer $g_j$ applies multi-head self-attention (MHSA) and a feed-forward network (FFN) with residual connections:
\begin{align}
\mathbf{H} &= \mathbf{X} + \mathrm{MHSA}(\mathrm{LN}(\mathbf{X})), \label{eq:mhsa} \\
\mathbf{Y} &= \mathbf{H} + \mathrm{FFN}(\mathrm{LN}(\mathbf{H})), \label{eq:ffn}
\end{align}
where $\mathbf{X}, \mathbf{Y} \in \mathbb{R}^{T \times d}$ are the input and output, $T$ is the sequence length, $d$ is the hidden dimension, and $\mathrm{LN}$ denotes layer normalization.
Following~\cite{wu-etal-2024-llama,cao2025progressive}, we zero-initialize the output projection matrices of both MHSA and FFN in the added layers, so that their outputs are zero at initialization.
Due to the residual connections and the absence of bias terms, $\mathbf{Y} = \mathbf{X}$ at initialization, making each added layer an identity function.
The added layers then gradually learn to adapt the representations for speech during training.

The position of added layers has been shown to affect performance in depth up-scaling for text domains~\cite{yano-etal-2025-step,cao2025progressive}.
We investigate this in the speech adaptation setting.
Following~\cite{wu-etal-2024-llama}, each added layer $g_j$ is initialized as a copy of the corresponding original layer $f_i$ and placed after it, forming the composition $g_j \circ f_i$.
We evaluate five placement strategies and compare them in our experiments:
\begin{itemize}
    \item \textsc{Interleaved}: distributes added layers $g$ evenly across all original layers $f$. 
    \item \textsc{Bottom}, \textsc{Middle}, and \textsc{Top}: concentrate added layers $g$ in the lower, middle, and upper half of the network, respectively.
    \item \textsc{Sandwich}: places added layers $g$ in both the bottom and top quarters.
\end{itemize}

\subsection{Added Layer Architecture}\label{sec:arch}

Prior work on depth up-scaling~\cite{wu-etal-2024-llama,yano-etal-2025-step,cao2025progressive} has used standard transformer layers as the added layers, identical to the original frozen layers.
However, since the added layers are responsible for adapting frozen representations to the acoustic modality, an architecture better suited for speech may be beneficial.

We therefore also consider E-Branchformer~\cite{kim2023branchformer}, which has been widely adopted in speech recognition~\cite{xu2025dynamic, peng2024owsm,wang2024mlca}, as an alternative architecture for the added layers.
While a standard transformer layer processes the input sequentially via self-attention and a feed-forward network, E-Branchformer processes the input through two parallel branches and combines them via a merge module.
Specifically, the global extractor branch applies MHSA to capture long-range dependencies, while the local extractor branch applies a convolutional spatial gating unit (CSGU) within a gated MLP (cgMLP) to capture local patterns:
\begin{align}
\mathbf{H}_G &= \mathrm{MHSA}(\mathrm{LN}(\mathbf{X})) \in \mathbb{R}^{T \times d}, \label{eq:global} \\
\mathbf{H}_L &= \mathrm{cgMLP}(\mathrm{LN}(\mathbf{X})) \in \mathbb{R}^{T \times d}. \label{eq:local}
\end{align}
The merge module then combines the two branch outputs by concatenation, applies a depthwise convolution ($\mathrm{DwConv}$) for spatial refinement, and projects back to $d$ dimensions:
\begin{align}
\mathbf{H}_{\mathrm{Merge}} &= (\mathbf{H}_C + \mathrm{DwConv}(\mathbf{H}_C))\, \mathbf{W}_{\mathrm{Merge}} \in \mathbb{R}^{T \times d}, \label{eq:merge}
\end{align}
where $\mathbf{H}_C = \mathrm{Concat}(\mathbf{H}_G, \mathbf{H}_L) \in \mathbb{R}^{T \times 2d}$ and $\mathbf{W}_{\mathrm{Merge}} \in \mathbb{R}^{2d \times d}$ is a linear projection.
The merge output $\mathbf{H}_{\mathrm{Merge}}$ is added to the input via a residual connection and then passed through a FFN as in Eq.~\eqref{eq:ffn}.\footnote{The original E-Branchformer uses macaron-style FFN (two half-step FFNs before and after the branches). We use a single FFN after the merge module for simplicity.}
Since the cgMLP branch is designed for acoustic processing, it is applied only to speech tokens during both training and inference, while text tokens bypass the cgMLP and merge module and use only the MHSA output.

For function preserving initialization of E-Branchformer layers, the zero-initialization strategy used for standard transformer layers (\S\ref{sec:depth}) cannot be directly applied.
In standard transformer layers, the output projection of MHSA ($\mathbf{W}^O \in \mathbb{R}^{d \times d}$) and FFN can be independently zero-initialized because each has its own residual connection.
In E-Branchformer, the two branches share a single merge projection $\mathbf{W}_{\mathrm{Merge}}$.
We initialize $\mathbf{W}_{\mathrm{Merge}} = [\mathbf{I}; \mathbf{0}]$, where $\mathbf{I} \in \mathbb{R}^{d \times d}$ is the identity matrix and $\mathbf{0} \in \mathbb{R}^{d \times d}$ is a zero matrix.
We also set $\mathbf{W}^O = \mathbf{0}$ and the output projection of the depthwise convolution to zero.
At initialization, the MHSA output is $\mathbf{H}_G = \mathbf{0}$ since $\mathbf{W}^O = \mathbf{0}$.
Therefore $\mathbf{H}_C \, \mathbf{W}_{\mathrm{Merge}} = [\mathbf{0}, \mathbf{H}_L] \cdot [\mathbf{I}; \mathbf{0}] = \mathbf{0}$, and the layer reduces to the identity $\mathbf{Y} = \mathbf{X}$, similar to that in Eqs.~\eqref{eq:mhsa} and~\eqref{eq:ffn}.
We evaluate the effectiveness of this initialization strategy in \S\ref{sec:results}.

\subsection{Layer Dropping at Inference}\label{sec:drop}

A direct benefit of freezing all original layers is that the added layers can simply be removed at inference time, analogous to stochastic depth~\cite{huang2016deep} and layer dropout~\cite{fan2020reducing}, recovering the original pre-trained model exactly and guaranteeing zero degradation on text tasks by design.
The full model is used when speech capability is needed, and the added layers are dropped when only text capability is required.

\begin{table}[t]
\caption{Results on SmolLM2-360M and SmolLM2-1.7B. WER (\%) on LibriSpeech (lower is better). Text task accuracy is the average of 8 benchmarks (\%, higher is better). $\Delta$ denotes the change from the pre-trained model. Depth up-scaling results are with added layers kept at inference. Dropping added layers recovers the pre-trained performance ($\Delta = 0.0$) by design. \textit{Italic} = full fine-tuning (updates all parameters). \textbf{Bold} = best among depth up-scaling.}
\label{tab:main}
\centering
\scalebox{0.75}{%
\setlength{\tabcolsep}{4pt}%
\begin{tabular}{@{}lllc cc cc@{}}
\toprule
\multirow{2}{*}{Model} & \multirow{2}{*}{Method} & \multirow{2}{*}{Position} & \multirow{2}{*}{Trainable} & \multicolumn{2}{c}{WER $\downarrow$} & \multicolumn{2}{c}{Text $\uparrow$} \\
\cmidrule(lr){5-6} \cmidrule(lr){7-8}
    & & & & clean & other & Avg & $\Delta$ \\
\midrule
\multirow{8}{*}{360M} & pre-trained & -- & -- & -- & -- & 56.8 & 0.0 \\
\cmidrule(lr){2-8}
    & Full fine-tuning & -- & 360M & \textit{2.7} & \textit{6.0} & 34.6 & -22.2 \\
    & LoRA & -- & 200M & 11.3 & 17.3 & 34.9 & -22.0 \\
\cmidrule(lr){2-8}
    & \multirow{5}{*}{Depth Up-scaling} & \textsc{Interleaved} & 200M & \textbf{3.1} & \textbf{6.7} & 54.9 & -1.9 \\
    & & \textsc{Sandwich} & 200M & 3.3 & 7.0 & 53.1 & -3.8 \\
    & & \textsc{Top} & 200M & 3.5 & 7.5 & 55.9 & \textbf{-0.9} \\
    & & \textsc{Middle} & 200M & 3.5 & 7.6 & 52.9 & -3.9 \\
    & & \textsc{Bottom} & 200M & 3.4 & 7.4& 51.0 & -5.8 \\
\midrule
\multirow{8}{*}{1.7B} & pre-trained & -- & -- & -- & -- & 69.0 & 0.0 \\
\cmidrule(lr){2-8}
    & Full fine-tuning & -- & 1.87B & \textit{2.3} & \textit{5.6} & 36.5 & -32.6 \\
    & LoRA & -- & 0.66B & 3.2 & 7.1 & 33.4 & -35.7 \\
\cmidrule(lr){2-8}
    & \multirow{5}{*}{Depth Up-scaling} & \textsc{Interleaved} & 0.66B & \textbf{2.4} & \textbf{5.5} & 60.8 & -8.3 \\
    & & \textsc{Sandwich} & 0.66B & 2.5 & 5.7 & 61.0 & -8.0 \\
    & & \textsc{Top} & 0.66B & 2.9 & 6.3 & 66.4 & \textbf{-2.7} \\
    & & \textsc{Middle} & 0.66B & 3.0 & 6.4 & 59.4 & -9.6 \\
    & & \textsc{Bottom} & 0.66B & 2.9 & 6.3 & 58.5 & -10.5 \\
\bottomrule
\end{tabular}}
\end{table}

\begin{table}[t]
\caption{Effect of added layer architecture (\textsc{Interleaved} placement) and function preserving initialization. Top: architecture comparison. Bottom: initialization comparison for E-Branchformer on SmolLM2-1.7B. $\mathbf{W}_{\mathrm{Merge}}$ is the merge projection (\S\ref{sec:arch}).}
\label{tab:arch}
\centering
\scalebox{0.80}{
\begin{tabular}{@{}llc cc cc@{}}
\toprule
\multirow{2}{*}{Model} & \multirow{2}{*}{Added layer type} & \multirow{2}{*}{Trainable} & \multicolumn{2}{c}{WER $\downarrow$} & \multicolumn{2}{c}{Text $\uparrow$} \\
\cmidrule(lr){4-5} \cmidrule(lr){6-7}
    & & & clean & other & Avg & $\Delta$ \\
\midrule
\multirow{3}{*}{360M} & Full fine-tuning & 360M & \textit{2.7} & \textit{6.0} & 34.6 & -22.2 \\
\cmidrule(lr){2-7}
    & Standard transformer & 200M & 3.1 & 6.7 & 54.9 & \textbf{-1.9} \\
    & E-Branchformer & 220M & \textbf{2.9} & \textbf{6.2} & 54.6 & -2.2 \\
\midrule
\multirow{3}{*}{1.7B} & Full fine-tuning & 1.87B & \textit{2.3} & \textit{5.6} & 36.5 & -32.6 \\
\cmidrule(lr){2-7}
    & Standard transformer & 0.66B & 2.4 & 5.5 & 60.8 & -8.3 \\
    & E-Branchformer & 0.75B & \textbf{2.3} & \textbf{5.3} & 62.3 & \textbf{-6.8} \\
\midrule
\multicolumn{7}{@{}l}{\textit{E-Branchformer initialization (1.7B)}} \\
\midrule
    & Random $\mathbf{W}_{\mathrm{Merge}}$ & 0.75B & 2.5 & 5.5 & 62.1 & -7.0 \\
    & Function preserving & 0.75B & \textbf{2.3} & \textbf{5.3} & \textbf{62.3} & \textbf{-6.8} \\
\bottomrule
\end{tabular}}
\end{table}

\begin{table*}[t]
\caption{Zero-shot prompt-based task outputs on SmolLM2-1.7B (\textsc{Interleaved}, E-Branchformer). Full FT = full fine-tuning, Depth = multimodal depth up-scaling. Input speech: ``\textit{A cold, lucid indifference reigned in his soul.}''}
\label{tab:qualitative}
\centering
\small
\begin{tabular}{@{}l p{3.2cm} p{4.5cm} p{4.5cm}@{}}
\toprule
Task & Prompt & Depth Up-scaling & Full FT \\
\midrule
Translation & Translate the sentence into French & Une indiff\'{e}rence froide, claire, r\'{e}gnait dans l'\^{a}me. & in in in in in in ... \\
Simplification & Rewrite into simple, modern English & A chilly indifference pervaded his soul. & in in in in in in ... \\
Summarization & Summarize in one clear sentence & The man was indifferent to everything. & in in in in in in ... \\
\bottomrule
\end{tabular}
\end{table*}

\section{Experiments}\label{sec:exp}

We evaluate the proposed multimodal depth up-scaling approach, comparing it against full fine-tuning and LoRA across two model scales, and further examine whether the retained text capabilities extend to zero-shot instruction following (\S\ref{sec:analysis}).

\subsection{Experimental Setup}\label{sec:exp_setup}

\textbf{Data.}
We use the English ASR subset of the OWSM v3.2 suite~\cite{tian2024effects} for training.
We focus exclusively on the ASR task and do not mix text-only data during training to evaluate text task preservation without the aid of text data replay\footnote{
Depth up-scaling is orthogonal to text data replay, and the two approaches can be combined. We intentionally omit text replay to isolate the effect of depth up-scaling on text task preservation.
}.

\noindent \textbf{Model.}
We adopt the SmolLM2 series~\cite{allal2025smollm} of sizes 360M (32 layers) and 1.7B (24 layers), pre-trained on up to 11T text tokens, as our base LLMs.
We follow OpusLM~\cite{tian25b_interspeech} for speech vocabulary expansion and tokenization.
For depth up-scaling (\S\ref{sec:depth}), we add $m$ transformer layers equal to 25\% of the original $n$ layers: $m = 8$ for the 360M model ($n = 32$, 0.20B added parameters, 0.51B total) and $m = 6$ for the 1.7B model ($n = 24$, 0.66B added parameters, 2.27B total). 
The added layers have the same hidden dimensions as the original model. 
We also test E-Branchformer~\cite{kim2023branchformer} layers (\S\ref{sec:arch}), which combine self-attention and convolution, as an alternative architecture for the added layers.

\noindent \textbf{Baselines.}
We compare against (1)~\textit{full fine-tuning}, which updates all parameters of the original model without adding layers, and (2)~\textit{LoRA}~\cite{hu2022lora}.
LoRA trains only low-rank adapters while the pre-trained weights remain frozen, and has been argued to be more robust against forgetting~\cite{biderman2024lora}.
The LoRA rank $r$ is set to match the trainable parameter count of depth up-scaling ($r = 144$ for 360M and $r = 356$ for 1.7B, yielding 0.20B and 0.66B trainable parameters, respectively)
\footnote{These ranks are much larger than typical LoRA configurations (e.g., $r = 16$), giving LoRA more capacity and making this a favorable setting for LoRA~\cite{biderman2024lora}.
}.

\noindent \textbf{Training.}
We train on approximately 48k hours of data using AdamW optimizer~\cite{loshchilov2018decoupled} with a peak learning rate of $1 \times 10^{-4}$, a 25k-step warmup, and linear decay to $2 \times 10^{-5}$, with a maximum context length of 8192 tokens, following~\cite{tian25b_interspeech}.

\noindent \textbf{Inference and Evaluation.}
We adopt a greedy search for ASR inference.
For ASR, we evaluate word error rate (WER) on the test-clean and test-other subsets of LibriSpeech~\cite{panayotov2015librispeech}.
For text-only capability, we evaluate eight benchmarks commonly used for pre-trained language model evaluation~\cite{yano2026pretraining}, namely ARC-Easy (ARC-e), ARC-Challenge (ARC-c)~\cite{clark2018think}, BoolQ~\cite{clark-etal-2019-boolq}, HellaSwag (HS)~\cite{zellers-etal-2019-hellaswag}, OpenBookQA (OBQA)~\cite{mihaylov-etal-2018-suit}, PIQA~\cite{piqa}, WinoGrande (WG)~\cite{sakaguchi2021winogrande}, and MMLU~\cite{hendrycks2021measuring}, using \texttt{olmes}~\cite{gu-etal-2025-olmes}.
We report the average text task score (Avg) and its change from the pre-trained model ($\Delta$).
For depth up-scaling, we report results with the added layers kept at inference to evaluate the text degradation, while dropping the added layers recovers the pre-trained model exactly by design as discussed in \S~\ref{sec:drop}.

\subsection{Results}\label{sec:results}

Table~\ref{tab:main} presents results for both model sizes, comparing full fine-tuning, LoRA, and the five depth up-scaling placement strategies described in \S\ref{sec:depth}.
The following trends are consistent across the two scales.

For ASR, depth up-scaling with \textsc{Interleaved} placement achieves WER comparable to full fine-tuning, while substantially outperforming LoRA.
LoRA performs particularly poorly on the smaller 360M model, suggesting that low-rank adaptation is insufficient when the base model has limited capacity.
Among placement strategies, \textsc{Interleaved} consistently achieves the best WER, followed by \textsc{Sandwich}, while \textsc{Top}, \textsc{Middle}, and \textsc{Bottom} show higher WER.
Comparing across scales, the WER gap between depth up-scaling (\textsc{Interleaved}) and full fine-tuning narrows as the model size increases, and on the 1.7B model, depth up-scaling matches or surpasses full fine-tuning on both test sets.

For text tasks, full fine-tuning and LoRA both cause severe degradation, while all depth up-scaling variants preserve text ability much better.
By design, dropping the added layers at inference (\S\ref{sec:drop}) recovers the original pre-trained model exactly ($\Delta = 0.0$), and even when keeping the added layers, degradation is far smaller than the baselines.
Notably, the ranking of placement strategies reverses between ASR and text preservation.
\textsc{Top} best preserves text performance but yields higher WER, while \textsc{Interleaved} achieves the best ASR but more text degradation, suggesting that distributing added layers throughout the network gives them more capacity to adapt representations for speech but also affects the original computation path more.

\subsection{Added Layer Architecture}\label{sec:arch_exp}

Using the \textsc{Interleaved} placement that achieved the best ASR among placement strategies (Table~\ref{tab:main}), we evaluate the effect of the added layer architecture proposed in \S\ref{sec:arch}.
Table~\ref{tab:arch} compares the standard transformer and E-Branchformer as the added layers.
For both model sizes, E-Branchformer improves WER over standard transformer layers.\footnote{E-Branchformer adds slightly more parameters due to the merge module and convolutional branch (Table~\ref{tab:arch}), but the improvement is consistent across both scales.}
On the 1.7B model, E-Branchformer added layers match or surpass full fine-tuning while retaining the text preservation advantage of depth up-scaling.
Text task preservation is comparable between the two architectures, indicating that the E-Branchformer substitution does not come at the cost of text performance.
These results suggest that the convolutional branch in E-Branchformer helps capture local acoustic patterns that complement the global context modeled by self-attention.

\subsubsection{Effect of Function Preserving Initialization}
Table~\ref{tab:arch} (bottom) compares the function preserving initialization described in \S\ref{sec:arch} against random $\mathbf{W}_{\mathrm{Merge}}$ initialization on SmolLM2-1.7B.
Function preserving initialization consistently outperforms random initialization, confirming that the $[\mathbf{I}; \mathbf{0}]$ identity mapping (\S\ref{sec:arch}), which preserves the pre-trained model's behavior at the beginning of training, leads to more effective adaptation.

\subsection{Qualitative Analysis: Retained Language Capabilities}
\label{sec:analysis}

As discussed in \S\ref{sec:drop}, removing the added layers fully restores the pre-trained model.
Here, we examine whether the model also preserves the base LLM's ability to follow text instructions when conditioned on spoken input, without using any explicit cascaded pipeline.
To this end, we evaluate zero-shot instruction-following beyond standard transcription by prepending a natural language instruction, together with a few in-context examples, to each spoken utterance, and asking the model to generate the target text directly from the speech-conditioned prompt.
Table~\ref{tab:qualitative} shows three such tasks: English-to-French translation, simplification, and summarization. Full fine-tuning fails completely, often producing degenerate outputs (e.g., repeated tokens).
In contrast, although trained exclusively on ASR data and without any text data replay, multimodal depth up-scaling produces coherent and task-appropriate outputs across all three settings through direct speech-conditioned text generation, consistent with the preserved text capabilities shown in Table~\ref{tab:main}.

\section{Conclusion}

We proposed multimodal depth up-scaling, where new transformer layers are inserted into a frozen text LLM and can be dropped at inference to fully recover the original text performance.
Experiments on SmolLM2-360M and SmolLM2-1.7B showed that depth up-scaling achieves ASR comparable to full fine-tuning while preserving text performance far better than both full fine-tuning and LoRA, and that E-Branchformer added layers further improve ASR.
Qualitative analysis confirmed that the model retains the base LLM's text instruction-following ability even without text data replay. Future work includes extending to other speech tasks and languages.

\clearpage

\section{Acknowledgement}
This work was supported by the JST Moonshot R\&D Grant Number JPMJMS2011-35 (fundamental research).
This work was partly achieved through the use of {the Supercomputer system or the Research cloud system or Interdisciplinary Large-scale Computing System} at the Information Initiative Center, Hokkaido University, Sapporo, Japan.

\bibliographystyle{IEEEtran}
\input{output.bbl}

\end{document}

%% file: output.bbl

%% file: main.bbl
\begin{thebibliography}{10}
\providecommand{\url}[1]{#1}
\csname url@samestyle\endcsname
\providecommand{\newblock}{\relax}
\providecommand{\bibinfo}[2]{#2}
\providecommand{\BIBentrySTDinterwordspacing}{\spaceskip=0pt\relax}
\providecommand{\BIBentryALTinterwordstretchfactor}{4}
\providecommand{\BIBentryALTinterwordspacing}{\spaceskip=\fontdimen2\font plus
\BIBentryALTinterwordstretchfactor\fontdimen3\font minus \fontdimen4\font\relax}
\providecommand{\BIBforeignlanguage}[2]{{%
\expandafter\ifx\csname l@#1\endcsname\relax
\typeout{** WARNING: IEEEtran.bst: No hyphenation pattern has been}%
\typeout{** loaded for the language `#1'. Using the pattern for}%
\typeout{** the default language instead.}%
\else
\language=\csname l@#1\endcsname
\fi
#2}}
\providecommand{\BIBdecl}{\relax}
\BIBdecl

\bibitem{brown2020language}
T.~Brown, B.~Mann, N.~Ryder, M.~Subbiah, J.~D. Kaplan, P.~Dhariwal, A.~Neelakantan, P.~Shyam, G.~Sastry, A.~Askell \emph{et~al.}, ``Language models are few-shot learners,'' \emph{Advances in neural information processing systems}, vol.~33, pp. 1877--1901, 2020.

\bibitem{grattafiori2024llama}
A.~Grattafiori, A.~Dubey, A.~Jauhri, A.~Pandey, A.~Kadian, A.~Al-Dahle, A.~Letman, A.~Mathur, A.~Schelten, A.~Vaughan \emph{et~al.}, ``The llama 3 herd of models,'' \emph{arXiv preprint arXiv:2407.21783}, 2024.

\bibitem{tang2024salmonn}
C.~Tang, W.~Yu, G.~Sun, X.~Chen, T.~Tan, W.~Li, L.~Lu, Z.~MA, and C.~Zhang, ``{SALMONN}: Towards generic hearing abilities for large language models,'' in \emph{The Twelfth International Conference on Learning Representations}, 2024.

\bibitem{chu2024qwen2}
Y.~Chu, J.~Xu, Q.~Yang, H.~Wei, X.~Wei, Z.~Guo, Y.~Leng, Y.~Lv, J.~He, J.~Lin \emph{et~al.}, ``Qwen2-audio technical report,'' \emph{arXiv preprint arXiv:2407.10759}, 2024.

\bibitem{tian25b_interspeech}
J.~Tian, W.~Chen, Y.~Peng, J.~Shi, S.~Arora, S.~Bharadwaj, T.~Maekaku, Y.~Shinohara, K.~Goto, X.~Yue, H.~Yang, and S.~Watanabe, ``{OpusLM: A Family of Open Unified Speech Language Models},'' in \emph{{Interspeech 2025}}, 2025, pp. 3259--3263.

\bibitem{arora2025on}
S.~Arora, K.-W. Chang, C.-M. Chien, Y.~Peng, H.~Wu, Y.~Adi, E.~Dupoux, H.~yi~Lee, K.~Livescu, and S.~Watanabe, ``On the landscape of spoken language models: A comprehensive survey,'' \emph{Transactions on Machine Learning Research}, 2025.

\bibitem{hsiao2025analyzing}
C.-Y. Hsiao, K.-H. Lu, K.-W. Chang, C.-K. Yang, W.-C. Chen, and H.~yi~Lee, ``{Analyzing Mitigation Strategies for Catastrophic Forgetting in End-to-End Training of Spoken Language Models},'' in \emph{{Interspeech 2025}}, 2025, pp. 3234--3238.

\bibitem{peng-etal-2025-voicetextblender}
Y.~Peng, K.~C. Puvvada, Z.~Chen, P.~Zelasko, H.~Huang, K.~Dhawan, K.~Hu, S.~Watanabe, J.~Balam, and B.~Ginsburg, ``{V}oice{T}ext{B}lender: Augmenting large language models with speech capabilities via single-stage joint speech-text supervised fine-tuning,'' in \emph{Proceedings of the 2025 Conference of the Nations of the Americas Chapter of the Association for Computational Linguistics: Human Language Technologies (Volume 1: Long Papers)}, L.~Chiruzzo, A.~Ritter, and L.~Wang, Eds.\hskip 1em plus 0.5em minus 0.4em\relax Albuquerque, New Mexico: Association for Computational Linguistics, Apr. 2025, pp. 5787--5802.

\bibitem{hu2022lora}
E.~J. Hu, yelong shen, P.~Wallis, Z.~Allen-Zhu, Y.~Li, S.~Wang, L.~Wang, and W.~Chen, ``Lo{RA}: Low-rank adaptation of large language models,'' in \emph{International Conference on Learning Representations}, 2022.

\bibitem{hu-etal-2024-wavllm}
S.~Hu, L.~Zhou, S.~Liu, S.~Chen, L.~Meng, H.~Hao, J.~Pan, X.~Liu, J.~Li, S.~Sivasankaran, L.~Liu, and F.~Wei, ``{W}av{LLM}: Towards robust and adaptive speech large language model,'' in \emph{Findings of the Association for Computational Linguistics: EMNLP 2024}, Y.~Al-Onaizan, M.~Bansal, and Y.-N. Chen, Eds.\hskip 1em plus 0.5em minus 0.4em\relax Miami, Florida, USA: Association for Computational Linguistics, Nov. 2024, pp. 4552--4572.

\bibitem{biderman2024lora}
D.~Biderman, J.~Portes, J.~J.~G. Ortiz, M.~Paul, P.~Greengard, C.~Jennings, D.~King, S.~Havens, V.~Chiley, J.~Frankle, C.~Blakeney, and J.~P. Cunningham, ``Lo{RA} learns less and forgets less,'' \emph{Transactions on Machine Learning Research}, 2024, featured Certification.

\bibitem{yang2025qwen3}
A.~Yang, A.~Li, B.~Yang, B.~Zhang, B.~Hui, B.~Zheng, B.~Yu, C.~Gao, C.~Huang, C.~Lv \emph{et~al.}, ``Qwen3 technical report,'' \emph{arXiv preprint arXiv:2505.09388}, 2025.

\bibitem{wu-etal-2024-llama}
C.~Wu, Y.~Gan, Y.~Ge, Z.~Lu, J.~Wang, Y.~Feng, Y.~Shan, and P.~Luo, ``{LL}a{MA} pro: Progressive {LL}a{MA} with block expansion,'' in \emph{Proceedings of the 62nd Annual Meeting of the Association for Computational Linguistics (Volume 1: Long Papers)}, L.-W. Ku, A.~Martins, and V.~Srikumar, Eds.\hskip 1em plus 0.5em minus 0.4em\relax Bangkok, Thailand: Association for Computational Linguistics, Aug. 2024, pp. 6518--6537.

\bibitem{yano-etal-2025-step}
K.~Yano, T.~Ito, and J.~Suzuki, ``{STEP}: Staged parameter-efficient pre-training for large language models,'' in \emph{Proceedings of the 2025 Conference of the Nations of the Americas Chapter of the Association for Computational Linguistics: Human Language Technologies (Volume 2: Short Papers)}, L.~Chiruzzo, A.~Ritter, and L.~Wang, Eds.\hskip 1em plus 0.5em minus 0.4em\relax Albuquerque, New Mexico: Association for Computational Linguistics, Apr. 2025, pp. 374--384.

\bibitem{cao2025progressive}
M.~Cao, X.~Wang, and N.~Aletras, ``Progressive depth up-scaling via optimal transport,'' \emph{arXiv preprint arXiv:2508.08011}, 2025.

\bibitem{allal2025smollm}
L.~B. allal, A.~Lozhkov, E.~Bakouch, G.~M. Blazquez, G.~Penedo, L.~Tunstall, A.~Marafioti, A.~P. Lajar{\'\i}n, H.~Kydl{\'\i}{\v{c}}ek, V.~Srivastav, J.~Lochner, C.~Fahlgren, X.~S. NGUYEN, B.~Burtenshaw, C.~Fourrier, H.~Zhao, H.~Larcher, M.~Morlon, C.~Zakka, C.~Raffel, L.~V. Werra, and T.~Wolf, ``Smol{LM}2: When smol goes big {\textemdash} data-centric training of a fully open small language model,'' in \emph{Second Conference on Language Modeling}, 2025.

\bibitem{tian2024effects}
J.~Tian, Y.~Peng, W.~Chen, K.~Choi, K.~Livescu, and S.~Watanabe, ``{On the Effects of Heterogeneous Data Sources on Speech-to-Text Foundation Models},'' in \emph{{Interspeech 2024}}, 2024, pp. 3959--3963.

\bibitem{kim2023branchformer}
K.~Kim, F.~Wu, Y.~Peng, J.~Pan, P.~Sridhar, K.~J. Han, and S.~Watanabe, ``E-branchformer: Branchformer with enhanced merging for speech recognition,'' in \emph{2022 IEEE Spoken Language Technology Workshop (SLT)}.\hskip 1em plus 0.5em minus 0.4em\relax IEEE, 2023, pp. 84--91.

\bibitem{wang2025continual}
G.~Wang, J.~Zhao, H.~Yang, G.~Qi, T.~Wu, and G.~Haffari, ``{Continual Speech Learning with Fused Speech Features},'' in \emph{{Interspeech 2025}}, 2025, pp. 1793--1797.

\bibitem{javed2025nirantar}
T.~Javed, K.~Bhogale, and M.~M. Khapra, ``{NIRANTAR: Continual Learning with New Languages and Domains on Real-world Speech Data},'' in \emph{{Interspeech 2025}}, 2025, pp. 918--922.

\bibitem{kim-etal-2024-solar}
S.~Kim, D.~Kim, C.~Park, W.~Lee, W.~Song, Y.~Kim, H.~Kim, Y.~Kim, H.~Lee, J.~Kim, C.~Ahn, S.~Yang, S.~Lee, H.~Park, G.~Gim, M.~Cha, H.~Lee, and S.~Kim, ``{SOLAR} 10.7{B}: Scaling large language models with simple yet effective depth up-scaling,'' in \emph{Proceedings of the 2024 Conference of the North American Chapter of the Association for Computational Linguistics: Human Language Technologies (Volume 6: Industry Track)}, Y.~Yang, A.~Davani, A.~Sil, and A.~Kumar, Eds.\hskip 1em plus 0.5em minus 0.4em\relax Mexico City, Mexico: Association for Computational Linguistics, Jun. 2024, pp. 23--35.

\bibitem{yang-etal-2025-lesa}
Y.~Yang, Z.~Cao, X.~Ma, Y.~Yao, Z.~Chen, L.~Qin, and H.~Zhao, ``{LESA}: Learnable {LLM} layer scaling-up,'' in \emph{Proceedings of the 63rd Annual Meeting of the Association for Computational Linguistics (Volume 1: Long Papers)}, W.~Che, J.~Nabende, E.~Shutova, and M.~T. Pilehvar, Eds.\hskip 1em plus 0.5em minus 0.4em\relax Vienna, Austria: Association for Computational Linguistics, Jul. 2025, pp. 22\,463--22\,476.

\bibitem{tian-etal-2025-espnet}
J.~Tian, J.~Shi, W.~Chen, S.~Arora, Y.~Masuyama, T.~Maekaku, Y.~Wu, J.~Peng, S.~Bharadwaj, Y.~Zhao, S.~Cornell, Y.~Peng, X.~Yue, C.-H.~H. Yang, G.~Neubig, and S.~Watanabe, ``{ESP}net-{S}peech{LM}: An open speech language model toolkit,'' in \emph{Proceedings of the 2025 Conference of the Nations of the Americas Chapter of the Association for Computational Linguistics: Human Language Technologies (System Demonstrations)}, N.~Dziri, S.~X. Ren, and S.~Diao, Eds.\hskip 1em plus 0.5em minus 0.4em\relax Albuquerque, New Mexico: Association for Computational Linguistics, Apr. 2025, pp. 116--124.

\bibitem{chen-etal-2022-bert2bert}
C.~Chen, Y.~Yin, L.~Shang, X.~Jiang, Y.~Qin, F.~Wang, Z.~Wang, X.~Chen, Z.~Liu, and Q.~Liu, ``bert2{BERT}: Towards reusable pretrained language models,'' in \emph{Proceedings of the 60th Annual Meeting of the Association for Computational Linguistics (Volume 1: Long Papers)}, S.~Muresan, P.~Nakov, and A.~Villavicencio, Eds.\hskip 1em plus 0.5em minus 0.4em\relax Dublin, Ireland: Association for Computational Linguistics, May 2022, pp. 2134--2148.

\bibitem{xu2025dynamic}
J.~Xu, Z.~Yang, A.~Zeyer, E.~Beck, R.~Schlüter, and H.~Ney, ``{Dynamic Acoustic Model Architecture Optimization in Training for ASR},'' in \emph{{Interspeech 2025}}, 2025, pp. 3603--3607.

\bibitem{peng2024owsm}
Y.~Peng, J.~Tian, W.~Chen, S.~Arora, B.~Yan, Y.~Sudo, M.~Shakeel, K.~Choi, J.~Shi, X.~Chang, J.~weon Jung, and S.~Watanabe, ``{OWSM v3.1: Better and Faster Open Whisper-Style Speech Models based on E-Branchformer},'' in \emph{{Interspeech 2024}}, 2024, pp. 352--356.

\bibitem{wang2024mlca}
H.~Wang, P.~Guo, P.~Zhou, and L.~Xie, ``Mlca-avsr: Multi-layer cross attention fusion based audio-visual speech recognition,'' in \emph{ICASSP 2024 - 2024 IEEE International Conference on Acoustics, Speech and Signal Processing (ICASSP)}, 2024, pp. 8150--8154.

\bibitem{huang2016deep}
G.~Huang, Y.~Sun, Z.~Liu, D.~Sedra, and K.~Q. Weinberger, ``Deep networks with stochastic depth,'' in \emph{European conference on computer vision}.\hskip 1em plus 0.5em minus 0.4em\relax Springer, 2016, pp. 646--661.

\bibitem{fan2020reducing}
\BIBentryALTinterwordspacing
A.~Fan, E.~Grave, and A.~Joulin, ``Reducing transformer depth on demand with structured dropout,'' in \emph{International Conference on Learning Representations}, 2020. [Online]. Available: \url{https://openreview.net/forum?id=SylO2yStDr}
\BIBentrySTDinterwordspacing

\bibitem{loshchilov2018decoupled}
I.~Loshchilov and F.~Hutter, ``Decoupled weight decay regularization,'' in \emph{International Conference on Learning Representations}, 2019.

\bibitem{panayotov2015librispeech}
V.~Panayotov, G.~Chen, D.~Povey, and S.~Khudanpur, ``Librispeech: An asr corpus based on public domain audio books,'' in \emph{2015 IEEE International Conference on Acoustics, Speech and Signal Processing (ICASSP)}, 2015, pp. 5206--5210.

\bibitem{yano2026pretraining}
K.~Yano, S.~Kiyono, S.~Kobayashi, S.~Takase, and J.~Suzuki, ``Pre-training {LLM} without learning rate decay enhances supervised fine-tuning,'' in \emph{The Fourteenth International Conference on Learning Representations}, 2026.

\bibitem{clark2018think}
P.~Clark, I.~Cowhey, O.~Etzioni, T.~Khot, A.~Sabharwal, C.~Schoenick, and O.~Tafjord, ``Think you have solved question answering? try arc, the ai2 reasoning challenge,'' 2018.

\bibitem{clark-etal-2019-boolq}
C.~Clark, K.~Lee, M.-W. Chang, T.~Kwiatkowski, M.~Collins, and K.~Toutanova, ``{B}ool{Q}: Exploring the surprising difficulty of natural yes/no questions,'' in \emph{Proceedings of the 2019 Conference of the North {A}merican Chapter of the Association for Computational Linguistics: Human Language Technologies, Volume 1 (Long and Short Papers)}, J.~Burstein, C.~Doran, and T.~Solorio, Eds.\hskip 1em plus 0.5em minus 0.4em\relax Minneapolis, Minnesota: Association for Computational Linguistics, Jun. 2019, pp. 2924--2936.

\bibitem{zellers-etal-2019-hellaswag}
R.~Zellers, A.~Holtzman, Y.~Bisk, A.~Farhadi, and Y.~Choi, ``{H}ella{S}wag: Can a machine really finish your sentence?'' in \emph{Proceedings of the 57th Annual Meeting of the Association for Computational Linguistics}, A.~Korhonen, D.~Traum, and L.~M{\`a}rquez, Eds.\hskip 1em plus 0.5em minus 0.4em\relax Florence, Italy: Association for Computational Linguistics, Jul. 2019, pp. 4791--4800.

\bibitem{mihaylov-etal-2018-suit}
T.~Mihaylov, P.~Clark, T.~Khot, and A.~Sabharwal, ``Can a suit of armor conduct electricity? a new dataset for open book question answering,'' in \emph{Proceedings of the 2018 Conference on Empirical Methods in Natural Language Processing}, E.~Riloff, D.~Chiang, J.~Hockenmaier, and J.~Tsujii, Eds.\hskip 1em plus 0.5em minus 0.4em\relax Brussels, Belgium: Association for Computational Linguistics, Oct.-Nov. 2018, pp. 2381--2391.

\bibitem{piqa}
Y.~Bisk, R.~Zellers, R.~{Le Bras}, J.~Gao, and Y.~Choi, ``Piqa: Reasoning about physical commonsense in natural language,'' \emph{Proceedings of the AAAI Conference on Artificial Intelligence}, vol.~34, pp. 7432--7439, 04 2020.

\bibitem{sakaguchi2021winogrande}
K.~Sakaguchi, R.~L. Bras, C.~Bhagavatula, and Y.~Choi, ``Winogrande: An adversarial winograd schema challenge at scale,'' \emph{Communications of the ACM}, vol.~64, no.~9, pp. 99--106, 2021.

\bibitem{hendrycks2021measuring}
D.~Hendrycks, C.~Burns, S.~Basart, A.~Zou, M.~Mazeika, D.~Song, and J.~Steinhardt, ``Measuring massive multitask language understanding,'' in \emph{International Conference on Learning Representations}, 2021.

\bibitem{gu-etal-2025-olmes}
Y.~Gu, O.~Tafjord, B.~Kuehl, D.~Haddad, J.~Dodge, and H.~Hajishirzi, ``{OLMES}: A standard for language model evaluations,'' in \emph{Findings of the Association for Computational Linguistics: NAACL 2025}, L.~Chiruzzo, A.~Ritter, and L.~Wang, Eds.\hskip 1em plus 0.5em minus 0.4em\relax Albuquerque, New Mexico: Association for Computational Linguistics, Apr. 2025, pp. 5020--5048.

\end{thebibliography}
